\documentclass[sigconf]{acmart}
\usepackage{float}

\usepackage{multirow}
\usepackage{xcolor}
\usepackage{setspace}

\setcopyright{none}
\copyrightyear{}
\acmYear{}
\acmDOI{}
\acmPrice{}
\acmISBN{}

\begin{document}

\title{Deriving Disinformation Insights \\ from Geolocalized Twitter Callouts}

\newcommand{\COMSCandCSRI}{
	\affiliation{%
		\department{Crime \& Security Research Institute}
		\department{School of Computer Science and Informatics}
		\institution{Cardiff University}
		\city{Cardiff}
		\country{UK}
  	}
  }

\author{David Tuxworth}
\COMSCandCSRI
\email{tuxworthdt@cardiff.ac.uk}
\author{Dimosthenis Antypas}
\COMSCandCSRI
\email{antypasd@cardiff.ac.uk}
\author{Luis Espinosa-Anke}
\COMSCandCSRI
\email{espinosa-ankel@cardiff.ac.uk}
\author{Jose Camacho-Collados} 
\COMSCandCSRI
\email{camachocolladosj@cardiff.ac.uk}
\author{Alun Preece}
\COMSCandCSRI
\email{preecead@cardiff.ac.uk}
\author{David Rogers}
\COMSCandCSRI
\email{rogersdm1@cardiff.ac.uk}

\acmConference[WIT2021]{KDD2021 Workshop On Deriving Insights From User-Generated Text}{August 14--18, 2021}{Singapore}
\renewcommand{\shortauthors}{Tuxworth and Antypas, et al.}

\begin{abstract}
This paper demonstrates a two-stage method for deriving insights from social media data relating to disinformation by applying a combination of geospatial classification and embedding-based language modelling across multiple languages. 
In particular, the analysis in centered on Twitter and disinformation for three European languages: English, French and Spanish. Firstly, Twitter data is classified into European and non-European sets using BERT.
Secondly, Word2vec is applied to the classified texts resulting in Eurocentric, non-Eurocentric and global representations of the data for the three target languages.
This comparative analysis demonstrates not only the efficacy of the classification method but also highlights geographic, temporal and linguistic differences in the disinformation-related media. Thus, the contributions of the work are threefold: (i) a novel language-independent transformer-based geolocation method; (ii) an analytical approach that exploits lexical specificity and word embeddings to interrogate user-generated content; and (iii) a dataset of 36 million disinformation related tweets in English, French and Spanish.
\end{abstract}

\maketitle

\section{Introduction}

Social media provides a rich stream of user-generated data that can be utilised in many ways. This paper employs a two-stage method to use this resource in order to derive insights into disinformation. The scale, immediacy and popularity of social media render it an ideal platform for the dissemination of ideas. While the many platforms available are used for legitimate communication, it is also used by modern propagandists to wilfully spread false information, i.e., disinformation. The inadvertent sharing of false information, i.e., misinformation, is widespread and while not necessarily malicious in intent, can be hugely damaging. Understanding the content targeted at as well as generated by users of social media is paramount in tackling these phenomena.  Computational methods are required not only to analyze but to keep pace with the volume of data generated by both legitimate and illegitimate users of social media. A further challenge is considering the language, culture and context of the messaging. These elements are considered in this paper.

The motivation for this study is practical, embedded in ongoing work to detect, track and understand disinformation operations in a variety of geopolitical contexts. To this end, Twitter data relating to misinformation, disinformation and related terms including propaganda and `fake news' have been continuously collected since 2019 in multiple languages including English, French and Spanish, which are the languages of focus in this study. The intuition behind the collection method is that Twitter users often `call out' misinformation and disinformation (following the definitions in \cite{shi2020}) through tagging or quoting media they find questionable. Of course, this does not mean that the media is actually misinformation or disinformation; often it is simply content that the users find objectionable. Nevertheless, collecting data with those terms (translated across the set of target languages) provides a superset of material for analysis. Given the global nature of English, French and Spanish, it becomes necessary to distinguish regional narratives, particularly the Americas versus Europe, from global ones. In turn, examining the use of language around specific query terms such as `immigrant'/`immigr\'e'/`inmigrante' can help derive insights into mis/disinformation narratives relating to those terms. How the use of language evolves over time is also potentially revealing.

To achieve this, the paper describes a two-stage method by which (1) user-generated data from Twitter is classified into European and non-European subsets in three languages: English, French and Spanish, and (2) embedding-based language models are built for each of the subsets, further subdivided into two periods of time. The choice of languages and time periods are illustrative; the method is completely general. English, French and Spanish were selected as a subset of languages for which data had been collected because all three are `global' languages relevant in the context of America and Europe. Time periods in 2019 and 2020 were selected because the former covered a period of significant political activity in Europe---the 9th European Parliament Elections, held during the time when the United Kingdom was in the process of leaving the European Union---and the latter covered the run-up to the 59th US Presidential Election; therefore, these two periods could be expected to provide distinctive regional narratives in each case. Moreover, the onset of the global Coronavirus pandemic in early 2020 would likely further differentiate narratives between the two periods, though with potentially less regional difference.

The main contributions of this work are (1) a novel transformer-based geolocation method that performs in multiple languages; and (2) an analytical method that uses lexical specificity and word embeddings to interrogate multilingual user-generated content with respect to mis/disinformation narratives. In addition, a dataset\footnote{\url{https://github.com/tuxworth/disinformation-insight-twitter}} of 36 million disinformation related tweets in English, French and Spanish is made available to researchers.

The paper is structured as follows:  Section~\ref{sec:related} summarises related work;
Section~\ref{sec:twitterdata} provides details of the multilingual disinformation-related dataset; 
\ref{sec:classification} presents the classification method and performance results;
\ref{sec:embeddings} describes the analytical method using lexical specificity and work embeddings; finally,
\ref{sec:conclusion} concludes the paper and highlights future work.

\section{Related Work}
\label{sec:related}

\subsection{Twitter Geolocalization}

Previous research \cite{BakermanJordan2018} shows that geotagging literature exists in three categories: network, text and hybrid methods.  A user's connections on social media are strong indicators of an individual's location \cite{jurgens2013s} and so it follows that network-based approaches have been highly successful in geolocalizing user locations. Work by \cite{compton2014geotagging} approaches the problem by inferring an unknown user's location through their friend's locations via a mention network. This technique is applied at scale in a distributed system enabling a predicted geolocation of millions of users. However, \citeauthor{huang2020} \cite{jurgens2015geolocation} have shown that exclusively network-based methods cannot geolocate all users, particularly those that do not form connections meaning there is no network structure available. 

The problem of geolocalizing non-geotagged tweets has been approached at varying levels of granularity including at the level of city neighborhoods by comparing the content of tweets to known geolocated examples \cite{ParaskevopoulosPavlos2016}. In this case the geographic regions, European or non-European, are far broader and are more comparable to country-level geolocation which has been shown to be a less challenging problem than city-level geolocation \cite{HuangBinxuan2017}.

A hybrid approach, combining both text and network features is recommended by \cite{jurgens2015geolocation} and \cite{BakermanJordan2018}. This is not possible in this case as the dataset excludes the attributes required to apply a network-based method and the tweet text is filtered by keywords resulting in the choice of employing metadata in the classification stage. It should be noted that the location and description are user-defined and are thus susceptible to data integrity issues whether by omission or using text which is not relevant or inaccurate. Despite this noise, experiments by \cite{HanBo2014} show that user-supplied locations contain valuable information and classifiers using the location field outperform purely text-based methods when predicting city-level location. In this work both user location and user description are leveraged as text features to feed into a machine learning classifier, using Twitter geolocalized tweets as seeds.
 
\subsection{Deriving Insights from Twitter Data}

Concerning the technical aspects relevant to this paper, this subsection focuses on the well-known word embeddings techniques and their applications to content analysis. The Word2vec \cite{mikolov2013a} toolkit, in its two variants CBOW and SkipGram, is one of the best known techniques for learning word embeddings. These dense vector representations have been leveraged extensively, for example, as input representations in neural network architectures for NLP tasks \cite{goldberg2017neural}, e.g., detecting `fake news' and phenomena related to the setting of this work \cite{thorne2018automated}. In a recent study identifying online propaganda \cite{KausarSoufia2020}, Word2vec embeddings were found to outperform a multilingual version of BERT in Urdu \cite{dong2019}, which the authors ascribe to the limited vocabulary of Urdu in the model. In another study, Word2vec has been leveraged as a feature in the detection of fake news where researchers found that it performs well in comparison to other textual features across multiple datasets and languages \cite{faustini2020}. Using ensemble methods to detect fake news, \cite{huang2020} use Word2vec as an embedding layer in a LSTM architecture.
	
In this paper, the learned embeddings are used to perform comparative analyses between classified sets of text rather than as input for a downstream task. The primary advantage of Word2vec is the ability to learn semantic relations between words via unsupervised machine learning. Word embedding models can be used to learn analogies (comparison between two elements based on limited shared characteristics). 	
	In fact, using word vector analogies as a proxy for understanding behaviours in online communities has been the focus, for example, in \cite{KobsKonstantin2020}, who used Twitch data to learn word and emoji embeddings which they then use to study Twitch-specific language, or in \cite{eisner2016emoji2vec}, who studied emoji analogies in Twitter-specific embeddings.  Finally, beyond analogies, Twitter embeddings have also been at the center of studies on gender and race \cite{barbieri2018gender}, as well as detecting semantic shift during the COVID-19 pandemic \cite{GuoYanzhu2021}.

\section{Twitter Data}
\label{sec:twitterdata}

\begin{figure}[h]
  \centering
  \includegraphics[width=\linewidth]{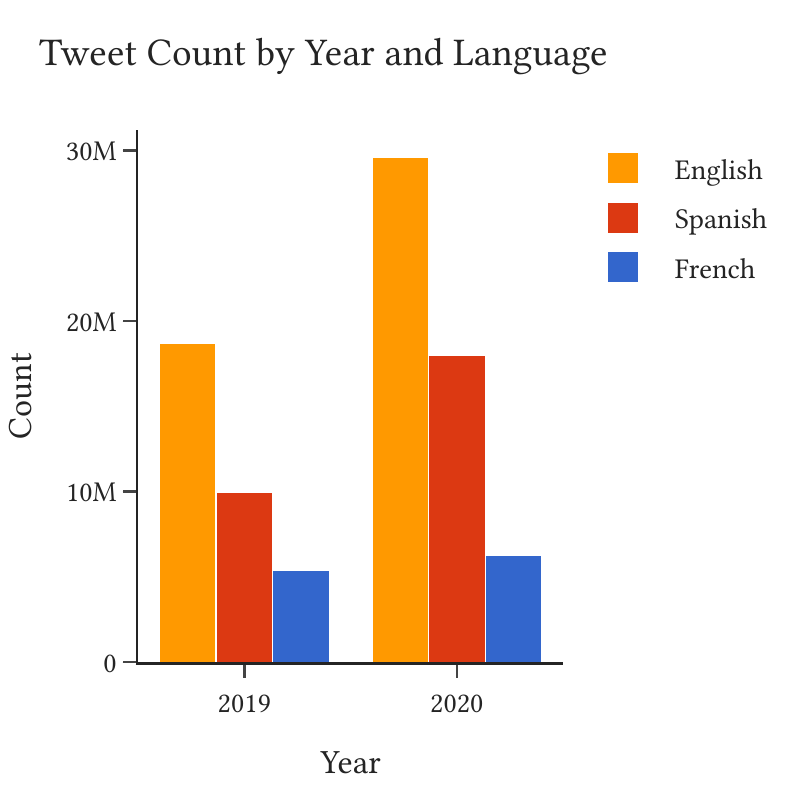}
  \caption{Number of tweets by year and language.}
  \Description{Number of tweets by year and language.}
  \label{figure:tweet_count}
\end{figure}

The data is collected via the Twitter API from two time periods: 2019-04-17 to 2019-06-30 and 2020-04-17 to 2020-6-30 (inclusive of both start and end dates). The 2019 range is selected as it covers the period surrounding the 2019 European Parliament elections that started the 23rd May. The 2020 range is selected to facilitate a comparative analysis between two years. 97 terms across the three languages were selected by subject-matter experts as being indicative of the concept of disinformation including: `misinformation', `fake news', `propaganda', and `lies'. These terms were used to collect the dataset. Three European languages are analyzed: English, Spanish and French selected by the `lang' attribute present within the tweet JSON. Figure \ref{figure:tweet_count} shows the proportion of tweets per language for the two years. The total number of tweets is 87,894,019 by 14,803,949 unique users.

294,877 tweets contain geolocation metadata which is 0.34\% of the total. To split the data into European and non-European tweets a classifier is trained using the samples that have geolocation data. The classifier is then applied to the remaining tweets that do not contain geolocation data. The class labels are derived from the country code. Tweets with geolocation metadata are labelled European if the country code matches one of those shown in Table \ref{table:country_codes} and non-European otherwise.

\begin{table}[h]
	\centering
	\caption{ISO 3166 country codes used to select the training data for the European class.}
	\resizebox{\columnwidth}{!}{
	\begin{tabular}{@{}c@{}}  \toprule
	ISO 3166 Country Code \\ \midrule
	AD, AL, AM, AT, AX, AZ, BA, BE, BG, BY, CH, CY, CZ, DE, DK, EE, ES, FI, FO, FR,\\
	GB, GE, GG, GI, GR, HR, HU, IE, IM, IS, IT, JE, KZ, LI, LT, LU, LV, MC, MD, ME,\\
	 MK, MT, NL, NO, PL, PT, RO, RS, RU, SE, SI, SK, SM, TR, UA, VA  \\ \bottomrule
	\end{tabular}}
	\label{table:country_codes}
\end{table}

\section{Geolocalization Classification}
\label{sec:classification}

As geolocation data is only available for 0.34\% of tweets, a method was developed to classify the data into geographic region. This section describes the methodology to attain location information for all tweets in the dataset.

\subsection{Experimental Setting}

\paragraph{Training and testing data} The subset of tweets which contain geolocation data from the full dataset are used to create a training corpus. Table \ref{table:train_datasets} shows the number of labelled tweets used for the geolocalization classification evaluation (all of them were subsequently used as training data to label the rest of the Twitter corpus). The user location and user description are used as features. For evaluation purposes a 80/10/10 (train/validation/test) stratified split is used for each language dataset. 

\begin{table}[h]
	\centering
	\caption{Distribution of tweets used for training the BERT classifiers.}
	\resizebox{\columnwidth}{!}{%
	\begin{tabular}{@{}cccccc@{}}  \toprule
		\multicolumn{3}{c}{2019}                   & \multicolumn{3}{c}{2020} \\ \cmidrule(lr){1-6}
		English		&Spanish		&French		&English		&Spanish		&French \\ \midrule
		60,430            	& 49,250            & 21,816           	& 74,206            & 66,820            & 22,355 \\ \bottomrule
	\end{tabular}}
	\label{table:train_datasets}
\end{table}

\paragraph{Preprocessing} A simple pre-processing step is applied to both the user description and user location where punctuation is removed and words (based on letters from the Unicode Basic Latin and Latin-1 Supplement) are extracted. User locations such as `New York' are concatenated to one term `new\_york'.

\paragraph{Text classification} Following this, a binary classifier is trained for each language using the user description and the user location as features and a Boolean label of `European' derived from the country code. Initially, a Naive Bayes classifier is used as a baseline model based on the implementation provided from scikit-learn \cite{scikit-learn}. Then, experiments are carried out with BERT-like models adapted for text classification. In total six models are trained and tested, one for each (language, year) combination.

\paragraph{Pre-trained language models} The BERT-base model \cite{devlin-etal-2019-bert} is used for the English language, while for Spanish and French BETO \cite{CaneteCFP2020} and  FlauBERT \cite{le2020flaubert} are applied respectively. All models trained are based on the implementations of the uncased versions provided by Hugging Face \cite{wolf-etal-2020-transformers}. Finally, we also experiment with a multilingual BERT model (mBERT).

\paragraph{BERT Optimization} All the BERT models were trained using the same process. Adam optimizer \cite{loshchilov2017decoupled} and a linear scheduler with warmup is utilized. We warm up linearly for 500 steps with a learning rate of $5\text{e-}5$, while a batch size n=34 is used. The models are trained up to 20 epochs, with a checkpoint in every epoch, while an early-stop callback stops the training process after 3 epochs without a performance increase of at least 0.01. We select the best model out of all the checkpoints based on their performance on the dev set.

\subsection{Results}

As Table \ref{table:bert-results} shows, the performance of the yearly BERT models is satisfactory for the task at hand with all the models achieving more than 85\% accuracy. For both 2019 and 2020 the English model appears to perform better (92\% F1-score)  while the French model produces the `worst' results with 87\% and 86\% F1-score. The difference in the performance could be justified by the smaller training datasets that were available for the Spanish and French languages (Table \ref{table:train_datasets}).

\begin{table*}[]
\caption{Classification results for the 2019 and 2020 datasets for each language model. Evaluation metrics: accuracy and macro-averaged precision, recall and F1. mBERT* model is trained on the whole corpus including the three languages. Naive baseline refers to a system where every tweet entry is classified as European}
\begin{tabular}{lllllllllllllll} \toprule 
		  		  &                                   &                                               & \multicolumn{4}{c}{English}                                    & \multicolumn{4}{c}{Spanish}                                    & \multicolumn{4}{c}{French} \\ \cmidrule(r){4-15}
                      Trained &                        Tested & Classifier			                                & Prec & Rec & Acc & F1   &  Prec & Rec & Acc & F1   & Prec & Rec & Acc & F1   \\ \midrule
			2019   &                          2019 & BERT                                                      & \textbf{0.94}      & \textbf{0.89}   & \textbf{0.95}              & \textbf{0.92} & \textbf{0.92}      & \textbf{0.85}   & \textbf{0.94}     & \textbf{0.88} & 0.89                & 0.86            & 0.9               & 0.87          \\ 
                                  &                                  & mBERT                                                     & 0.93               & \textbf{0.89}   & \textbf{0.95}     & 0.91          & 0.91               & 0.84   & 0.93              & 0.87          & \textbf{0.91}       & \textbf{0.87}   & \textbf{0.91}     & \textbf{0.89} \\ 
                                  &                                  & mBERT*                                                    & \textbf{0.94}      & 0.86            & 0.94              & 0.89          & 0.51               & 0.51            & 0.74              & 0.51          & 0.48                & 0.49            & 0.64              & 0.47          \\ 
                                  &                                  & Naive Bayes                                               & 0.89               & 0.81            & 0.92              & 0.84          & 0.88               & 0.81            & 0.92              & 0.84          & 0.86                & 0.81            & 0.87              & 0.83          \\ \addlinespace
			2020   &             	      2020 & BERT                                                      & \textbf{0.95}      & \textbf{0.88}   & \textbf{0.96}     & \textbf{0.92} & \textbf{0.94}      & 0.84            & 0.94              & 0.88          & \textbf{0.91}       & \textbf{0.84}   & 0.89              & \textbf{0.86} \\ 
                                  &                                  & mBERT                                                     & \textbf{0.95}      & 0.86            & 0.95              & 0.9           & \textbf{0.94}      & \textbf{0.85}   & \textbf{0.95}     & \textbf{0.89} & 0.9                 & 0.83   & \textbf{0.9}      & \textbf{0.86} \\ 
                                  &                                  & mBERT*                                                    & 0.94               & 0.87            & 0.95              & 0.9           & 0.5                & 0.5             & 0.74              & 0.5           & 0.49                & 0.5             & 0.63              & 0.47          \\ 
                                  &                                  & Naive Bayes                                               & 0.9                & 0.82            & 0.92              & 0.85          & 0.9                & 0.82            & 0.93              & 0.85          & 0.86                & 0.83            & 0.87              & 0.84          \\ 
			2019   & 2020                          & BERT                                                      & 0.94               & 0.89            & 0.95              & 0.91          & 0.92               & 0.85            & 0.94              & 0.88          & 0.91                & 0.87            & 0.91              & 0.89          \\  \addlinespace
&                                & Naive Baseline                                           				     & 0.25               & 0.5             & 0.5               & 0.33          & 0.25               & 0.5             & 0.5               & 0.33          & 0.25                & 0.5             & 0.5               & 0.33          \\ \bottomrule
\end{tabular}
\label{table:bert-results}
\end{table*}

\paragraph{Cross-temporal analysis} An effort was made to train and use BERT models only using the 2019 data. The classification metrics when tested on the 2020 data (Table \ref{table:bert-results}: \textit{Bert 2019/2020}), indicate that even though for the Spanish and French datasets the model's performance is on par (same F1 score for Spanish) or even slightly better for French with the models trained on each year, the performance on the English dataset drops (from 92\% to 91\% F1 score). This shows that BERT classifiers based on user descriptions are robust even for different periods from where it was trained, which can be relevant for practical settings.

\paragraph{Multilingual BERT} A multilingual BERT model (mBERT) is trained and tested using the combined language datasets for 2019 and for 2020. Unfortunately, training on all languages did not lead to improvements and indeed the results were inferior (see Table \ref{table:bert-results}: \textit{mBERT*}).
However, the same multilingual model is competitive for all languages when trained on individual language datasets separately.
In this case there is an improved performance on the French dataset for 2019 (87\% to 89\% F1 score) and on the Spanish dataset for 2020 (88\% to 89\% F1 score) when compared to individual models.
\newline

Most of the models trained displayed similar performances when tested. It is possible that by using a different multilingual implementation, or further fine-tuning the existing multilingual model, better results could be achieved compared with using monolingual models across all languages. At the same time, it has been observed in related research \cite{wu2020all}\cite{wu2019beto} that for high resources languages, like the ones investigated, mBERT can perform worse than monolingual BERT models depending the task. As the main objective was inferring the location of unseen tweets it was decided to use different models for each language for each year studied. The monolingual BERT model indeed achieved the best results for the largest part of our corpus (English tweets subset). The selected monolingual BERT classifiers are then applied to the rest of the data to create the European and non-European sets. This enables us to analyze the Twitter corpus collected as described in Section~\ref{sec:twitterdata}, with all tweets tagged with location information.

\section{Analysis}
\label{sec:embeddings}

\begin{figure}[h]
  \centering
  \includegraphics[width=\linewidth]{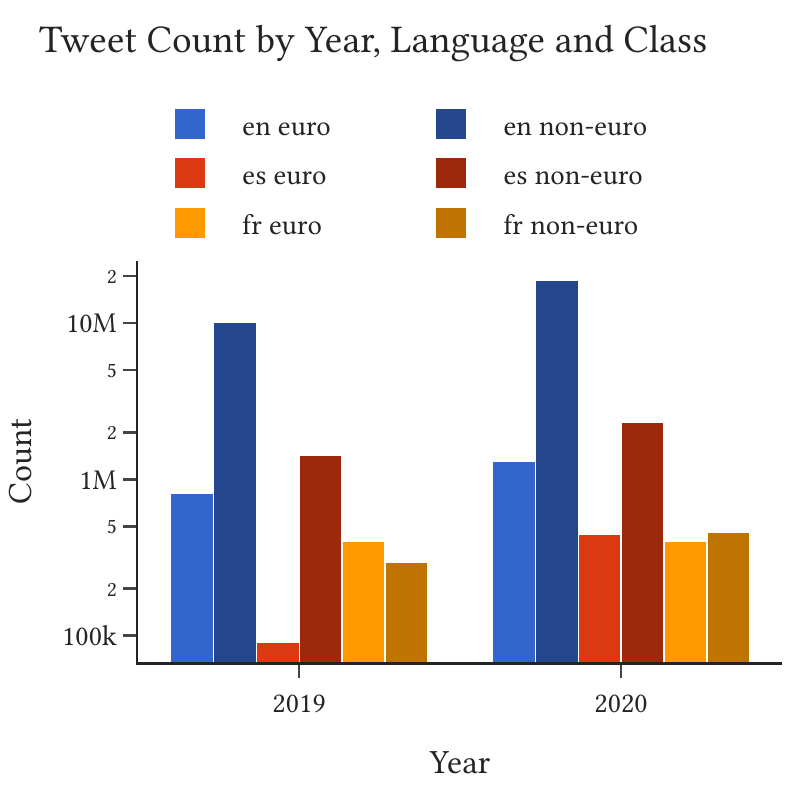}
  \caption{Filtered Tweet Count by Year, Language and Class.}
  \Description{Filtered Tweet Count by Year, Language and Class.}
  \label{figure:tweet_count_filtered}
\end{figure}

To enable a balanced comparison between languages, the classified tweet texts are filtered to include only those that match a subset of terms originally used to collect the data. The terms, shown in Table \ref{table:filter-terms}, revolve around disinformation, propaganda and themes of influence. Figure \ref{figure:tweet_count_filtered} shows the classified tweets after applying this step. The total number of tweets is 36,655,061.

\begin{table}[ht]
	\caption{Terms used to filter tweets for training the embeddings by each language.}
	\begin{tabular}{p{2.3cm}p{2.3cm}p{2.3cm} }   \toprule
	        English &         Spanish &          French \\ \midrule
	active measures &medidas activas & mesures actives \\
	     conspiracy &conspiración&complot \\
	        deceive &engañar&tromper \\
	     deep state&estado profundo &état profond \\
	 disinformation&desinformación&désinformation \\
	    fabrication&invención &invention \\
	      fake news &noticias falsas&fausse nouvelle \\
	      influence &influencia&influence \\
	   interference &interferencia &ingérence \\
	     manipulate &manipular & manipuler \\
	 misinformation & desinformación & désinformation \\
	     propaganda & propaganda & propagande \\
	     subversion &subversión &subversion \\ \bottomrule
	\end{tabular}
	\label{table:filter-terms}
\end{table}

The following section describes analyses to derive insights from this geolocalized corpus of tweets, by means of lexical specificity (Section \ref{sec:le}) and word embeddings (Section \ref{sec:emb}).

\subsection{Lexical Specificity}
\label{sec:le}
Initially, an attempt was made to identify similarities and differences between the European and non-European tweets for each language subset. This was achieved by computing the lexical specificity value of each word. Lexical specificity is a statistical measure which calculates the set of most representative words for a given text based on a reference corpus and the hypergeometric distribution \cite{lafon1980variabilite,camacho2016nasari}. In contrast to similar scores used to calculate importance of terms, such as TF-IDF, lexical specificity is not especially sensitive to different text lengths and does not require a full partition of the corpus.

\begin{table*}[h]
\caption{Top terms, along with their respective lexical specificity score, for the European (E) and non-European (NE) subsets of each language for each year studied.}
\begin{tabular}{lllllll}
\toprule
\multirow{6}{*}{2019} & English E  &      brexit - 17569    &      die - 14801   &     bbc - 14330             &   electoral - 9389                 &   farage - 5883           \\                                        
                      & English NE &      trump - 19487             &    mueller - 9453               &   obama - 7935               &     media - 7216               &      president - 7067              \\
                      & Spanish E    & advertencia - 235   & esbirros - 176  &  hecha - 146 &  terrorista - 130 & asesina - 122  \\      
                      & Spanish NE & banco - 204          & engañar - 164         & presidente - 137      & quer - 121 & bolsonaro - 109       \\
                      & French E   & faire - 2837 &   plus - 2355  &   fait - 2282   &  macron - 1415       &  monde - 1245         \\ 
                      & French NE  & mueller - 1050  & trump - 1039  & faux- 724   & clinton - 501 & spécial - 492          \\ \addlinespace
\multirow{6}{*}{2020} & English E  &       tory -  12389            &     boris - 11499          &   cummings - 10597               &       forgotten - 9098            &     johnson -  8879                \\
                      & English NE &   trump - 13024                &      president - 8497             &   obama - 7107               &    democrats - 4807                &    election  - 3944                  \\
                      & Spanish E  & sánchez - 5093     & sono - 4463    &   españa - 3484    & gobierno - 3211   &    vox - 2931     \\ 
                      & Spanish NE & trump - 7791     & india - 3444  & fox - 2457 &  ccp - 2286 & própria - 2258            \\ 
                      & French E   & plus - 2232  & faire - 2226    & fait - 1879      & meuf - 1639       & bien - 1348  \\ 
                      & French NE  & eua - 1388   & sedition - 1224  & secession - 1219  & ccp - 754  &   venezuela - 639           \\ \bottomrule
\end{tabular}
\label{table:lexical_specificity}
\end{table*}

Table \ref{table:lexical_specificity} displays, for each language, the top five relevant terms according to lexical specificity with respect to the corpus of each year, when considering the European and non-European subsets separately. To gain a better understanding of tweets content, Table \ref{table:lexical_specificity} does not include words that do not belong to the respective language (e.g. only French words were considered for the French subsets). One interesting observation is that for every language the European and non-European sets appear to have different terms. For example, for the English 2019 subset the European corpus is focused on the topic of Brexit while in the non-European corpus terms were found related to USA politics (e.g., `trump' and `obama'). Similarly, when considering the Spanish 2020 subset the European part revolves around Spain with terms like `sánchez' (Pedro Sánchez being the Spanish prime minister) and `españa', while the non-European subset seems to be more international with terms like `ccp', `india' and `trump'. These results verify, in a way, that the classification process applied was successful.

Another interesting observation is the almost complete change of topic for the English European corpus from Brexit related terms in 2019 to more generic political ones in 2020. There is also an evolution of the Spanish European corpus from intimidating terms in 2019, such as `terrorista' (terrorist) and `esbirros' (thugs), to a more `nationalistic' turn in 2020 with terms like `españa' (Spain) and `gobierno' (government).

\subsection{Embeddings}
\label{sec:emb}
The natural language processing libraries spaCy \cite{spacy2020} and gensim \cite{gensim2010} are used to preprocess the tweet texts. The extended version of the tweet is used and retweets are included. The text is tokenized and lemmatized with punctuation removed. The `RT' token present at the start of any retweets as well as any urls are removed. The phrase detection technique introduced by \citeauthor{mikolov2013b} \cite{mikolov2013b} is applied to the text with significant bigrams concatenated into a single string delimited by an underscore character. These phrases are considered individual tokens in training.

While pre-trained models have become the foundation to many NLP applications, they are primarily designed to generalize. In this case the latent aspects of interest can be more easily discovered by training a language model using solely the data to be investigated. To achieve this, Word2vec \cite{mikolov2013a} is used with the continuous bag-of-words (CBOW) model architecture to create the embeddings. 


\begin{table*}[t]
    \centering
    \caption{The 10 most similar words to the query by year and geographic region for English.}
	\begin{tabular}{@{}lllllll@{}} \toprule
	&\multicolumn{3}{l}{2019 English}&\multicolumn{3}{l}{2020 English} \\ \cmidrule(r){2-7}
        Query&All&European& Non-European&All&European&Non-European\\ \midrule
         immigrant & migrant & migrant & immigrants & immigrants & refugee & immigrants \\
	         & immigrants & semites & immigration & immigration & migrant & immigration \\
	         & immigration & immigration & migrant & foreigner & foreigner & foreigner \\
	         & refugee & zionists & jews & refugee & migrants & deportation \\
	         & jews & refugee & blacks & latinos & greeks & refugee \\
	         & mexicans & musli & mexicans & mexicans & europeans & blacks \\
	         & quidproquo & suffragette & refugee & asians & pensioner & mexicans \\
	         & blacks & jews & invader & invader & settlement & latinos \\
	         & invader & vaxer & quidproquo & latino & libyans & asians \\
	         & emigrant & semite & labourer & deportation & asians & latino \\
        
         vaccine & vaccination & vaccination & vaccination & vaccination & vaccination & vaccination \\
	         & vaxxers & vape & vaccinations & vaccines & vaccines & vaccines \\
	         & vaxer & measles & vaxx & vacine & malaria & vacine \\
	         & vaccinations & vaxxers & vaxxers & mmr & cure & mmr \\
	         & vape & vaccines & vaxer & medication & tetanus & vac \\
	         & vaxx & measle & vape & vac & microchip & rubella \\
	         & vaccineswork & tesla & vaxxe & immunization & mmr & medication \\
	         & measle & vaxer & measle & microchip & rfid & cure \\
	         & vaxxe & mmr & vaccinateyourkids & microchippe & jab & microchip \\
	         & vaccinateyourkids & virus & vac & cure & patent & vaxxe \\ \bottomrule
    \end{tabular}
    \label{table:english-results}
\end{table*}

\paragraph{English} Table \ref{table:english-results} shows the ten most similar words for two queries, `immigrant' and `vaccine' for each year and by geographic region in English. For the `immigrant' query the most striking result is the learned terms for ethnic groups that would be expected to be associated with the geographic region. For example `greeks' and `europeans' in the 2020 English European model compared with `mexicans' and `blacks' in the English non-European model. There are expected terms mixed in as well such as `immigration', `migrant', `refugee' and `foreigner'. Other differences include multiple  learned terms relating to Judaism (`jews', `zionists', `semites') in the 2019 European English set which are not present in the 2020 European set indicating a shift in the topics. These examples show a clear difference in the use of the word in and outside of Europe  in the context of disinformation.

There are also notable differences for the query `vaccine', particularly to do with conspiracy theories. One of the most popular conspiracies was the assertion that the 2020 Coronavirus Pandemic was a ruse to inject microchips via vaccines. As can be seen in the 2020 English results, `microchip' and `rfid' feature in the most similar words to vaccine showing that this method has the ability to identify emerging or trending conspiracies. 


\begin{table*}[t]
	\centering
	\caption{The 10 most similar words to the query by year and geographic region for Spanish.}
	\begin{tabular}{@{}lllllll@{}} \toprule
		&\multicolumn{3}{l}{2019 Spanish}&\multicolumn{3}{l}{2020 Spanish} \\ \cmidrule(r){2-7}
		Query&All&European& Non-European&All&European&Non-European\\ \midrule
        inmigrante & perjuicio & perjuicio & perjuicio & copia & copia & copia \\
	         & embajada & laicidad & inmigracion & mapuch & televisión\_sectario & mapuch \\ 
	         & inmigracion & divisa & adve & inmigración & derribo & turista \\ 
	         & laicidad & via & embajada & turista & example & inmigración \\ 
	         & amanecerrcn & prado\_miembro & renta & colono & estratagema & example \\ 
	         & backstage & estados & leyva & etnia & sodomía & vivienda \\ 
	         & demócrata & años & cuneta & paguita & fachada & campesino \\ 
	         & republicanos & cultivo & etnia & islam & difamación & sirios \\ 
	         & etnia & inspección & rebelión & gitanos & niña & crer \\ 
	         & manada & descarga & estancamiento & inmigracion & acoso & beneficios \\ 
        vacuna & vih & vih & vph & vacunación & chip & tratamiento \\ 
	         & vph & live & vih & tratamiento & laboratorio & vacunación \\ 
	         & neumonia & investigación & vacunación & cura & microchip & cura \\ 
	         & anorexia & mod & neumonia & medicamento & medicamento & medicamento \\ 
	         & estigma & auge & gripe & vacunas & bill\_gates & microchip \\ 
	         & leaving\_neverland & taller & mkt & chip & virus & gripe \\
	         & inmunización & ataque & fármaco & microchip & nanochip & virus \\
	         & pornografía & irak & virus & virus & sida & inyección \\
	         & pastilla & acciones & inmunización & sida & vacunación & vacunas \\
	         & vacunación & phishing & musicoterapia & vih & humanidad & chip \\ \bottomrule
	\end{tabular}
	 \label{table:spanish-results}
\end{table*}

\begin{table*}[t]
	\centering
	\caption{The 10 most similar words to the query by year and geographic region for French.}
	\begin{tabular}{@{}lllllll@{}} \toprule
	&\multicolumn{3}{l}{2019 French}&\multicolumn{3}{l}{2020 French} \\ \cmidrule(r){2-7}
	Query&All&European& Non-European&All&European&Non-European\\  \midrule
        immigré & invasion & souche & athmane\_tartag & délinquance & colonie & immigration \\
	        & arabie\_saoudite & pauvreté & mohamed\_médiène & colonie & tradition & réfugié \\
	        & civil & république & moise & immigration & banlieue & banlieue \\
	        & pauvreté & humiliation & abdallah & réfugié & délinquance & ouest \\
	        & occupation & résistance & fisc & banlieue & algérie & esclavagisme \\
	        & quota & président & nezzar & paysan & souveraineté & occupation \\
	        & référendum & travailleur & macky & monarchie & souverain & colonie \\
	        & nation & nationalisme & triade & tribu & richesse & délinquance \\
	        & résistance & richesse & glyphosate\_monsanto & tradition & colonisation & tradition \\
	        & traître & invasion & impérialisme & esclavagisme & esclavage & terrorisme \\ 
        vaccin & vaccination & vaccination & veritable\_islam & remède & vaccination & traitement \\
	        & glyphosate & lutte & lutte & médicament & puce & médicament \\
	        & lutte & généralisation & glorieuse\_nation & vaccination & id2020 & remède \\
	        & méfiance & maladie & noms & puce & chloroquine & puce \\
	        & ameriquelatine & élevage & triade & médoc & médoc & virus \\
	        & élevage & mobutu & antisémitisme & chloroquine & médicament & hcq \\
	        & blanchiment & polio & signataire & bill\_gate & bill\_gate & big\_pharma \\
	        & scrat & mutinerie & diatlov & traitement & traitement & bill\_gates \\
	        & maçonnerie & eglise & populisme & id2020 & hcq & vaccination \\
	        & généralisation & lyme & élevage & gates & big\_pharma & covid \\ \bottomrule
    \end{tabular}
    \label{table:french-results}
\end{table*}

\paragraph{Spanish} Table \ref{table:spanish-results} shows the ten most similar words for two queries, `inmigrante' (immigrant) and `vacuna' (vaccine) for each year and by geographic region in Spanish. 
For the query `inmigrante' (immigrant) the most similar word across all three geographic regions for 2019 is `perjuicio' (damage/detriment) which suggests that the word is being used in a negative context.  For 2020, the top word across all three geographic regions  is `copia' (copy)  which initially appears odd. However, on inspecting the data there are multiple retweets about creating a propaganda video for Vox (a far-right Spanish political party) blaming immigrants for selling pirated media.

For the query `vacuna' (vaccine) there is a clear difference between the two years. The top results 2019  include `vih' (HIV) and `vph' (HPV) which mirror common misinformation and disinformation spread by anti-vaxxer groups stating that vaccines result in these illnesses. There are also words that would be expected such as `inmunización' (immunization), `vacunación' (vaccination) and gripe (flu) as well as unexpected words such as `pornografía' (pornography) and `irak' (Iraq). For the year 2020, there are results more in keeping with what would be expected from a generalized language model mixed in with multiple terms
to do with consiracy theories such as `microchip' and `bill\_gates'. One of the most popular conspiracies was the assertion that the 2020 Coronavirus Pandemic was a ruse to inject microchips via vaccines.


\paragraph{French} Table \ref{table:french-results} shows the ten most similar words for two queries, `immigré' (immigrant) and `vaccin' (vaccine) for each year and by geographic region in French. For the query `immigré' (immigrant) the most similar terms for non-European 2019 are `athmane\_tartag' and `mohamed\_médiène` referring to the arrest of two Alegerian intelligence officials. The rest of the results for 2019 are quite mixed with many of the words being related to ideologies or pertain to the ruling of the state for example `nationalisme' (nationalism), `république' (republic) and `nation' (nation).
For both years there are terms that suggest a threat such as `occupation' (occupation), `invasion' (invasion) and terrorisme (terrorism) which is language common in far-right rhetoric.

For the query `vaccin' (vaccine) `big\_pharma' appears in reference to a conspiracy theory that states the pharmaceutical industry has malevolent ulterior motives. This is especially relevant as the period is in the beginnings of the 2020 COVID-19 pandemic. `id2020' is a genuine organisation that provides identification services. Misinformation spread stating that a vaccination program by the organisation and Bill Gates aimed to give people worldwide a digital ID. `Hydroxycholorquine' and an abbreviation `hcq' refer to the antimalarial medicine that misinformation categorised as a `cure' for Coronavirus when in reality it was an experimental treatment.

\subsubsection{Analogical Reasoning} 

One of the main benefits of word embeddings, as shown in \cite{mikolov2013a,mikolov2013b} is the ability to perform analogical reasoning by computing the relational similarity between two word pairs $\left<\mathbf{a},\mathbf{b}\right>$ and $\left<\mathbf{c},\mathbf{d}\right>$ by finding the most similar word associated with the resulting vector $\mathbf{d}$ (measured usually by cosine distance) to a query consisting on $\mathbf{d}=\mathbf{b}+\textbf{a}-\textbf{c}$. For example `London - Britain + Spain = Madrid', which in natural language can be phrased as  `London is to Britain as Madrid is to Spain'. In such case, a \textit{capital-of} relationship is learned and revealed via this operation. Table \ref{table:analogy2019} and Table \ref{table:analogy2020} list examples of these arithmetic operations using the English embeddings we use in this paper. The third element on each row is predicted by using the first, second and fourth words.

The first row shows that the `American' and `British' qualities of media organizations have been learned with different outlets for 2019 and 2020, `abc' and `fox' respectively. In the second row for 2020, the learned analogy is incorrect with `drumpf' being the original surname of Donald Trump's family. The third row shows a more generic example, with different short forms for `doctor'.

\begin{table}
\caption{Analogical reasoning examples using the English 2019 All Word2vec model (predictions in bold).}
\begin{tabular}{llll} \toprule	  
          \multicolumn{4}{l}{English 2019 All} \\ \midrule
          bbc &    britain &           \textbf{abc} &  america \\
        trump &    america & \textbf{boris\_johnson} &  britain \\
	politician & government &      \textbf{md} & hospital \\ \bottomrule
\end{tabular}
 \label{table:analogy2019}
\end{table}

\begin{table}
	   \caption{Analogical reasoning examples using the English 2020 All Word2vec model (predictions in bold).}
\begin{tabular}{llll} \toprule
           \multicolumn{4}{l}{English 2020 All} \\ \midrule
          bbc &    britain &      \textbf{fox} &  america \\
        trump &    america & \textbf{drumpf} &  britain \\
   politician & government &      \textbf{drs} & hospital \\ \bottomrule
\end{tabular}
 \label{table:analogy2020}
\end{table}

\subsubsection{Disinformation Surrounding the Origins of COVID-19}

A particularly successful conspiracy in the English language from early 2020 was that COVID-19 originated in a laboratory. Various flavours of this disinformation circulated ranging from rumours that the virus had been accidentally released to assertions that it was an American or Chinese biological weapon. Table \ref{table:laboratory-results} shows the top 5 most similar words to `laboratory' in the 2019 and 2020 English All models. There is a clear absence of terms relating to this conspiracy in 2019 and the strong presence of it in 2020. Other conspiratorial themes appear in the French and Spanish embeddings models though these are omitted for brevity.

The most similar words for 2019 are mundane terms that are related to the word `laboratory'. In comparison with 2020, the most similar words relate to this conspiracy including `wuhan' and `wuhan\_lab' for the Wuhan Institute of Virology, and the United States military lab `fort\_detrick' for the American version. These relate to the United States and Chinese counterparts of these analogous strands of disinformation. These examples show a dramatic change in the use of the term `laboratory' in the context of disinformation. This finding aligns with other studies, which have used word embeddings to demonstrate semantic shift during the pandemic \cite{GuoYanzhu2021}.

\begin{table}[h]
	\caption{The 5 most similar words to the query `laboratory' by year for the model English All.}
	\begin{tabular}{ll}
	\toprule
	English All 2019 & English All 2020 \\
	
	\midrule
	furniture &              lab \\
	warehouses &           biolab \\
	rebar &        wuhan\_lab \\
	extrusion &     fort\_detrick \\
	shoplife &            wuhan \\
	\bottomrule
	\end{tabular}
	 \label{table:laboratory-results}
\end{table}


\section{Conclusion \& Future Work}
\label{sec:conclusion}

This paper shows that user-generated content in multiple languages can be used as a data source for deriving insights into disinformation. To achieve this, first a transformer-based classifier is trained on the 0.34\% of 87.9 million tweets that contain geolocation data which is then applied to the rest of the data, separating it into European and non-European tweets. This is done for two periods, 2019 and 2020, in English, French and Spanish allowing for multiple types of comparative analysis. It is demonstrated that monolingual classifiers trained and tested on data from the same year outperform multilingual classifiers. Furthermore, it is shown that the geolocation metadata from a relatively small subset of tweets can be used to classify the entire set. An advantage of this method is that the data used to train the classifier is self-contained and usable so long as there is a large enough volume of geolocated tweets to make machine learning methods viable. Secondly, lexical specificity and word embeddings are used to explore the classified tweets and reveal insights into disinformation. For example, it is shown that the conspiracies surrounding the origin of COVID-19 are revealed through comparing the most similar words to a relevant keyword.

Future work could include classifying the data at a lower levels of granularity, for instance at country level by simply using the country code instead of grouping them into broader regions. A popular method of visualising word embeddings is by projecting the vectors into 2 dimensions using a method such as t-SNE \cite{van2008visualizing}. This type of visualisation could form part of an end-to-end system that would allow subject-matter experts with limited technical training to conduct these analyses. Experiments are also being conducted to turn the results of the analytic methods into query and `dashboard' tools for analysts. 

\bibliographystyle{ACM-Reference-Format}
\bibliography{bibliography}

\end{document}